\documentclass{article}
\usepackage{spconf,amsmath,graphicx}
\usepackage{verbatim}

\usepackage{amsfonts,amssymb}
\usepackage{booktabs}
\usepackage{multirow}
\usepackage{caption}

\title{Text Region Multiple Information Perception Network for Scene Text Detection}
\name{Jinzhi Zheng$^{1,2}$ \qquad Libo Zhang$^{1,2} \sthanks{Corresponding author: libo@iscas.ac.cn}$ \qquad Yanjun Wu$^{1}$  \qquad  Chen Zhao$^{1}$}
\address{
$^{1}$ Institute of Software Chinese Academy of Sciences, Beijing, China \\
$^{2}$ University of Chinese Academy of Sciences, Beijing, China\\
}
\begin{document}
%
\maketitle
\begin{abstract}
Segmentation-based scene text detection algorithms can handle arbitrary shape scene texts and have strong robustness and adaptability, so it has attracted wide attention. Existing segmentation-based scene text detection algorithms usually only segment the pixels in the center region of the text, while ignoring other information of the text region, such as edge information, distance information, etc., thus limiting the detection accuracy of the algorithm for scene text. This paper proposes a plug-and-play module called the Region Multiple Information Perception Module (RMIPM) to enhance the detection performance of segmentation-based algorithms. Specifically, we design an improved module that can perceive various types of information about scene text regions, such as text foreground classification maps, distance maps, direction maps, etc. 
Experiments on MSRA-TD500 and TotalText datasets show that our method achieves comparable performance with current state-of-the-art algorithms.
\end{abstract}
\section{Introduction}
\label{sec:intro}

Scene text detection is the task of detecting text in scene images, which is one of the important contents of scene understanding and has significant implications for artificial intelligence \cite{1}. Due to its wide application value in areas such as autonomous driving, image retrieval, product recommendation, visual dialogue, scene analysis, etc., it has attracted widespread attention \cite{2,3,8,33}. Especially with the development of artificial intelligence\cite{31,32}, scene text detection algorithms have made significant progress. However, due to the complexity and diversity of natural scenes, such as the arbitrary shapes, variable sizes, aspect ratios, and random distribution of text positions in scenes, current scene text detection algorithms still face many challenges \cite{1,4,29}.

Scene text detection algorithms based on deep learning can usually be divided into regression-based  algorithms\cite{5,6} and segmentation-based algorithms \cite{3,4} . In the early stages, deep learning-based algorithms treated scene text detection as a task similar to general object detection and located the text regions by regressing the bounding boxes\cite{9,5,7}. For example, EAST \cite{9} locates text regions by regressing text boxe and using quadrilaterals with rotation angles to represent the detected text regions. TextBoxes \cite{5} adopts an anchor-based detection method for text detection, 
while TextBoxes++ \cite{7} enables the detection of oriented text by designing offset values for quadrilateral text boxes. 
Regression-based detection methods can adapt to regular horizontal or vertical text detection. However, irregular texts are more common in scenes, so segmentation-based methods for detecting scene text have been proposed. For example, DB++ \cite{4} segments the central region of the text and expands it outward to obtain the complete text region. Segmentation-based algorithms have pixel-level accuracy and can adapt to arbitrary shape scene text, making them suitable for detecting irregular scene texts. However, current segmentation-based algorithms usually only segment the information of the text center region, ignoring other information of the text regions, which limits the performance of such algorithms.

\begin{figure*}[t]
\centering
\includegraphics[width=0.9\textwidth]{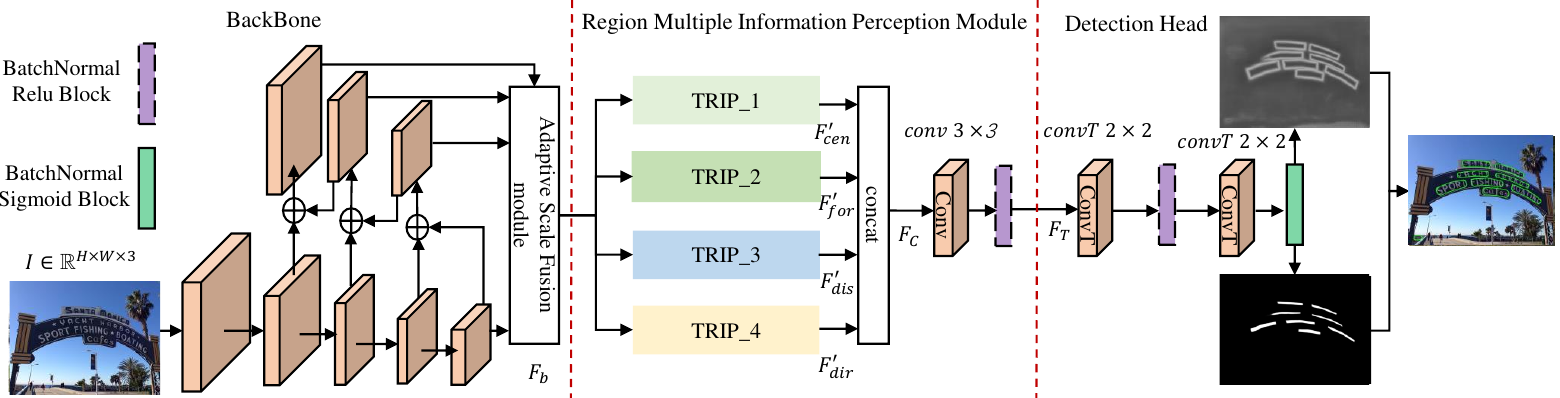}
\caption{The overall architecture of our RMIPN, which mainly consists of a Backbone, a Region Multiple Information Perception Module (RMIPM), and a Detection Head.  
}
\label{Figure1.}
\end{figure*}

Based on the above analysis, to enable segmentation-based scene text detection 
algorithms to perceive text region multiple information, we propose a plug-and-play module called the Region Multiple Information Perception module (RMIPM), which enhances the detection accuracy of scene text. Through the designed RMIPM, various types of information about text regions
can be added during the training process. 
The main contributions of this paper can be summarized as follows: 
1) We propose a plug-and-play module called  Region Multiple Information Perception Module (RMIPM), which enables scene text detection algorithms to perceive text region multiple information during the feature extraction process. 2) We embed the RMIPM  into the baseline model to propose the Region Multiple Information Perception Network(RMIPN), which can perceive various regional information according to different perceptual targets, such as center information, foreground classification map, pixel distance map, pixel direction map, etc. 3) Extensive experiments show that the proposed algorithm achieves competitive performance, demonstrating the robustness and effectiveness of the proposed approach.

\section{PROPOSED METHOD}
\label{sec:format}

\subsection{Overview}
\label{sec:Overview}
We used DBNet\cite{3} as the baseline, and the RMIPN is illustrated in Fig.\ref{Figure1.}. Given an input image $I\in \mathbb{R}^{H\times W \times 3}$ of scene text to be detected, the backbone network extracts text visual features $F_b\in \mathbb{R}^{\frac{H}{4}\times\frac{W}{4}\times C}$ at different scales. Then, RMIPM perceives various types of information about the text region to obtain text region features. Finally, the detection head completes text detection based on the text region features.
\subsection{Detailed Architecture}
\label{sec:Detailed Architecture}
\noindent {\bfseries Backbone:} 
U-Net has been widely used in visual tasks. In order to make a fair comparison with previous algorithms, we use the U-Net with a pyramid structure backbone similar to DB++ \cite{4} as the model backbone. Specifically, given the input image $I\in \mathbb{R}^{H\times W \times 3}$ to be detected, Resnet extracts visual features as follows:
\begin{equation}
F_b=U\left(I\right)\in \mathbb{R}^{\frac{H}{4}\times\frac{W}{4}\times C}
\end{equation}
where $H$ and $W$ represent the height and width of the original image, $C$ represents the feature dimension, and $U$ represents U-Net with ASF module\cite{4}. It is also possible to replace U-Net as a backbone with another network, such as a feature pyramid \cite{28}.

\noindent {\bfseries RMIPM:} Existing segmentation-based scene text detection algorithms usually only perform pixel-level segmentation on the text center region, while ignoring other text region information. In order to enable segmentation-based algorithms to perceive multiple information about text regions, we have proposed a plug-and-play module called Region Multiple Information Perception Module (RMIPM). The RMIPM is composed of multiple sub-modules called Text Region Information Perception Modules (IPM).

In this paper, we designed a text center map, a foreground classification map, a distance map from the pixels in the text area to the text edges, and a direction map of the pixels in the text region pointing to the text edges. The process of text region information perception by the IPM can be formulated as follows:
\begin{equation}
\left\{\begin{array}{l}
F_C=cat(F_{cen}^\prime, F_{for}^\prime, F_{dis}^\prime, F_{dir}^\prime),
\\F_T=R(Conv_{3\times3}(F_C)\in \mathbb{R}^{\frac{H}{4}\times\frac{W}{4}\times C}
\end{array}
\right.
\end{equation}
where $F_{cen}^\prime$, $F_{for}^\prime$, $F_{dis}^\prime$, and $F_{dir}^\prime$ represent the perceived center region information, foreground classification information, distance information from the pixels in the text area to the text edges, and direction information of the pixels in the text regions pointing to the text edges. $Conv_{3\times3}$ represents a convolution layer with a $3\times3$ kernel, $R$ represents the $ReLU$ activation functions.

\begin{figure}[t]
\centering
\includegraphics[width=0.45\textwidth]{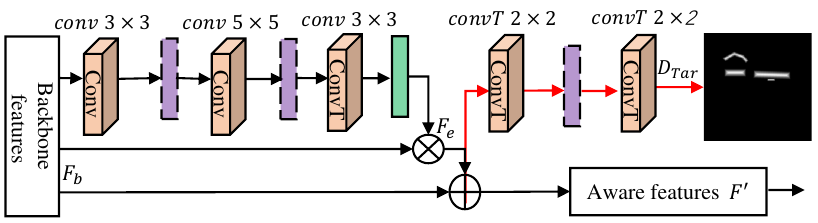}
\caption{The overall architecture of Text Region Information Perception Module (IPM). Red arrows indicate that they are only present during the training phase}
\label{Figure2.}
\end{figure}

\noindent {\bfseries IPM:} In order to perceive various types of information in the text region, we design the text region information perception module(IPM). The structure of the text region information perception module is shown in Fig.\ref{Figure2.}. The feature of text region information perception can be formulated as follows:

\begin{equation}
\left\{\begin{array}{l}
W_a=S(Con{vT}_{3\times3}(R(Conv_{5\times5}(R(Conv_{3\times3}(F_b)))))),
\\F_e=W_a\times F_b\in \mathbb{R}^{\frac{H}{4}\times\frac{W}{4}\times C},
\\F^\prime=F_e+F_b 
\end{array}
\right.
\end{equation}
where $Conv_{5\times5}$ represents a convolution layer with a $5\times5$ kernel, $Con{vT}_{3\times3}$ represents a  deconvolution layer with a $3\times3$ kernel, $S$ represents the $Sigmoid$ functions.

In the process of supervision training, the text region information perception goal can be formulated as:
\begin{equation}
D_{Tar}=\ Con{vT}_{3\times3}(R(Con{vT}_{3\times3}(F_e))))\in \mathbb{R}^{H\times W \times n}
\end{equation}
where $n$ is the channel of the feature corresponding to the perceived text region. For example, in this paper, four types of targets are set. When the perceived region information is a direction map, $n$ takes the value of 2 (Including horizontal and vertical directions). Otherwise, $n$ takes the value of 1.

\noindent {\bfseries Detection Head:} 
After obtaining the perceived text region features $F_T$, the detection head completes the text detection. We also used a commonly used convolution structure \cite{4}, and the detection head can be formulated as follows:
\begin{equation}
D_T=S(Con{vT}_{2\times2}(R(Con{vT}_{2\times2}(F_T)))
\end{equation}
where $Con{vT}_{2\times2}$ represents a deconvolution layer with a $2\times2$ kernel and batch normalization.

\begin{center}
\begin{figure}[h]
\begin{minipage}[c]{0.32\linewidth}
\centerline{\includegraphics[width=\textwidth,height=0.80in]{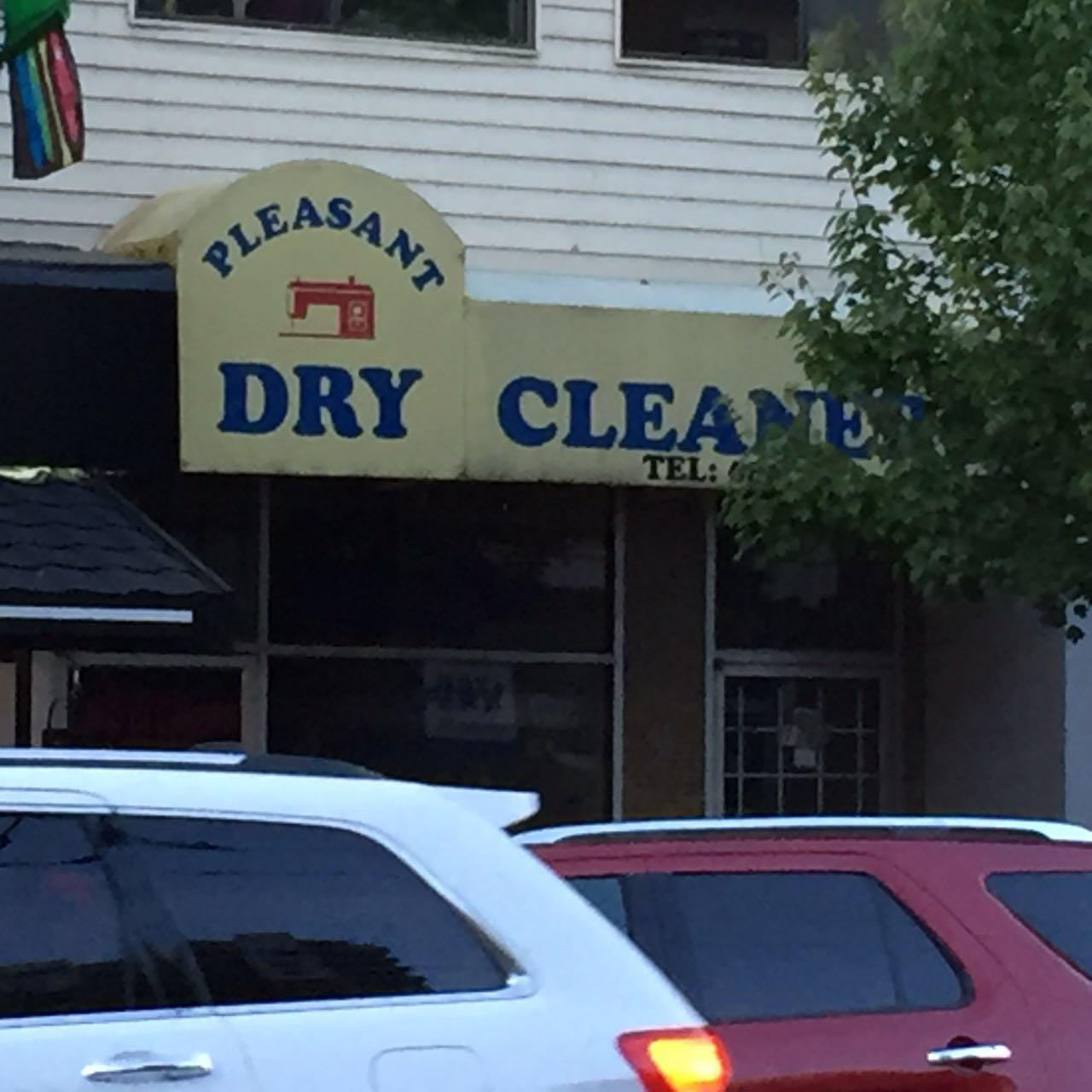}}
\centerline{{(a) Sample Image}}
\vspace{3pt}
\centerline{\includegraphics[width=\textwidth,height=0.8in]{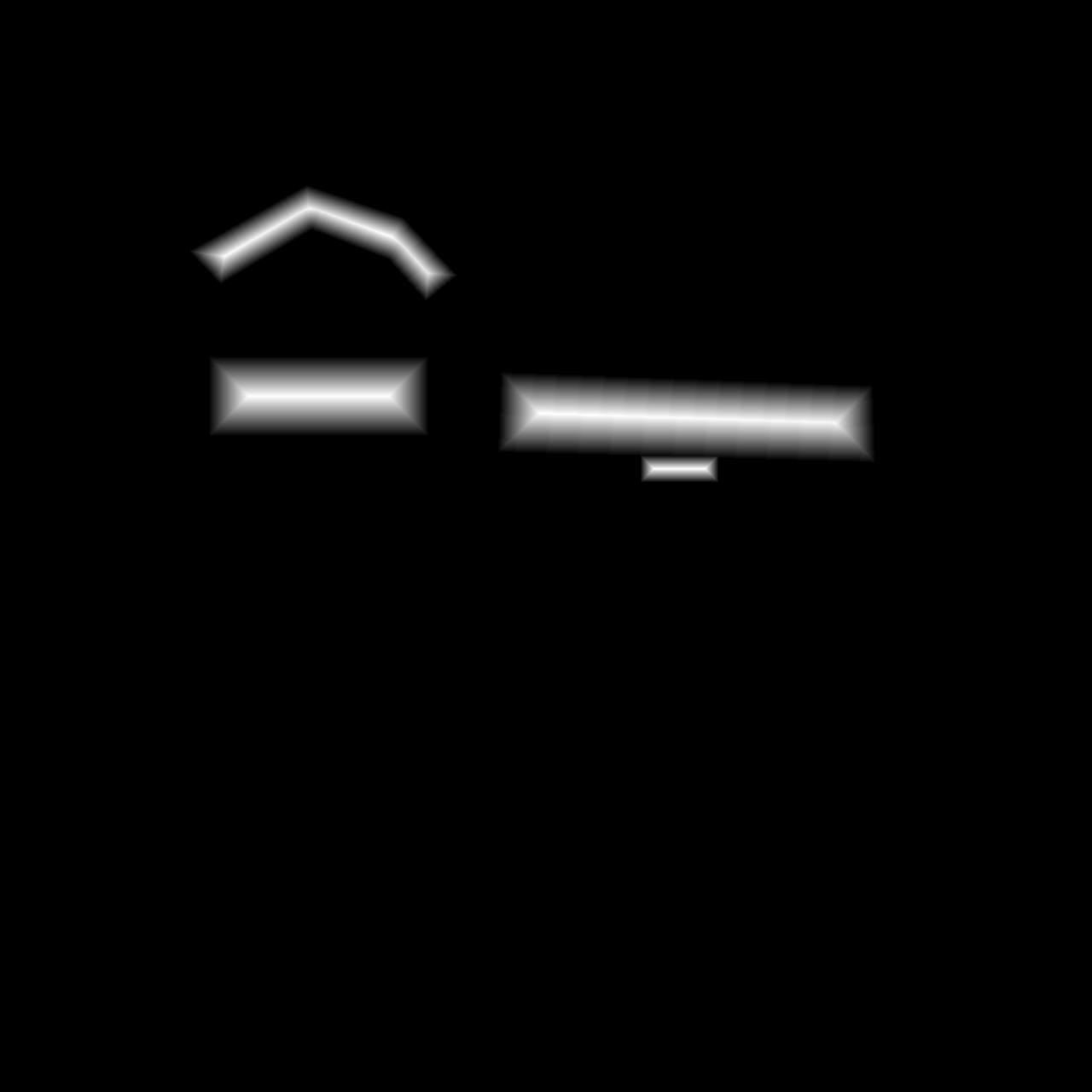}}
\centerline{{(d) Distance Map}}
\end{minipage}
\vspace{3pt}
\begin{minipage}[c]{0.32\linewidth}
\centerline{\includegraphics[width=\textwidth,height=0.8in]{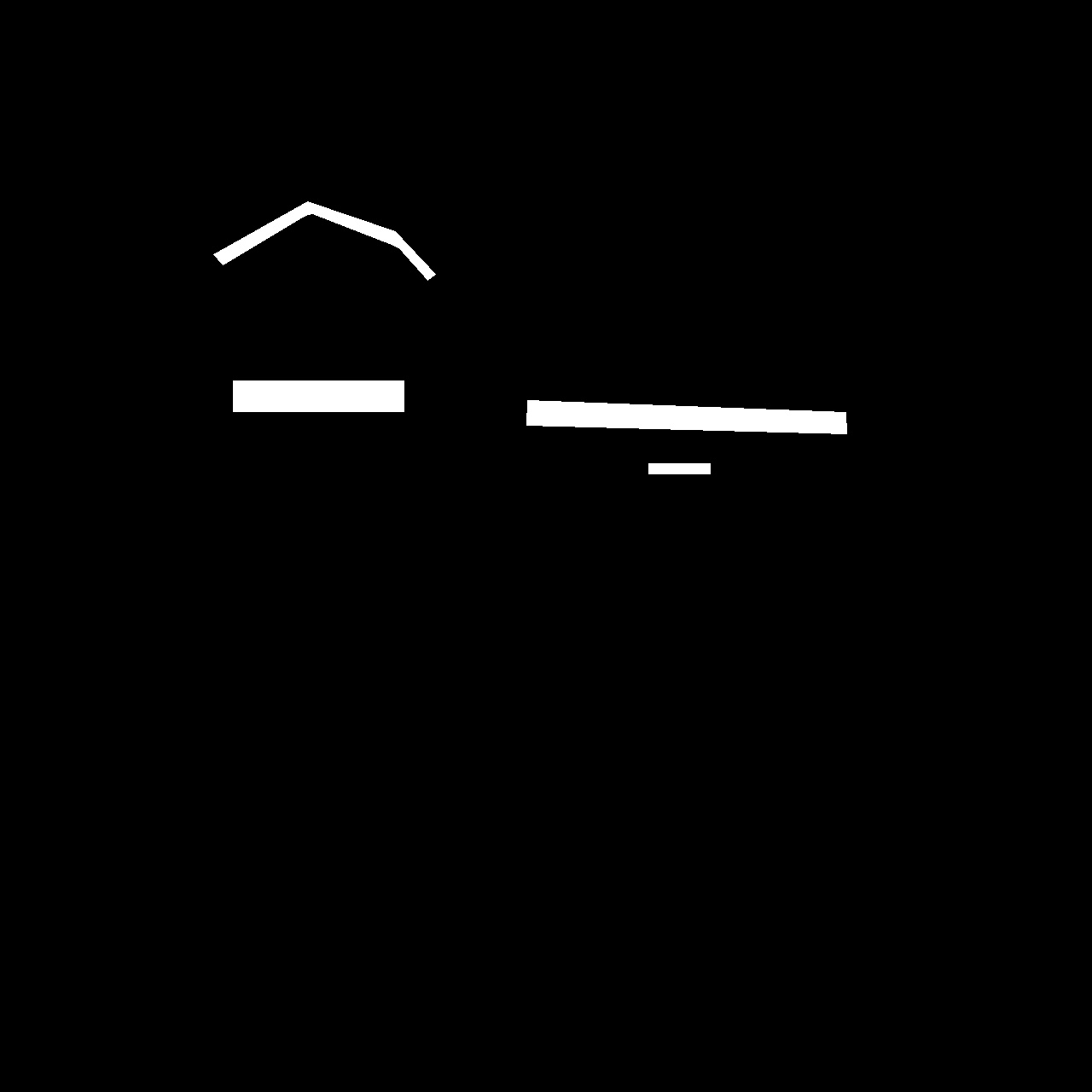}}
\centerline{\small{(b) Text Center Map}}
\vspace{3pt}
\centerline{\includegraphics[width=\textwidth,height=0.8in]{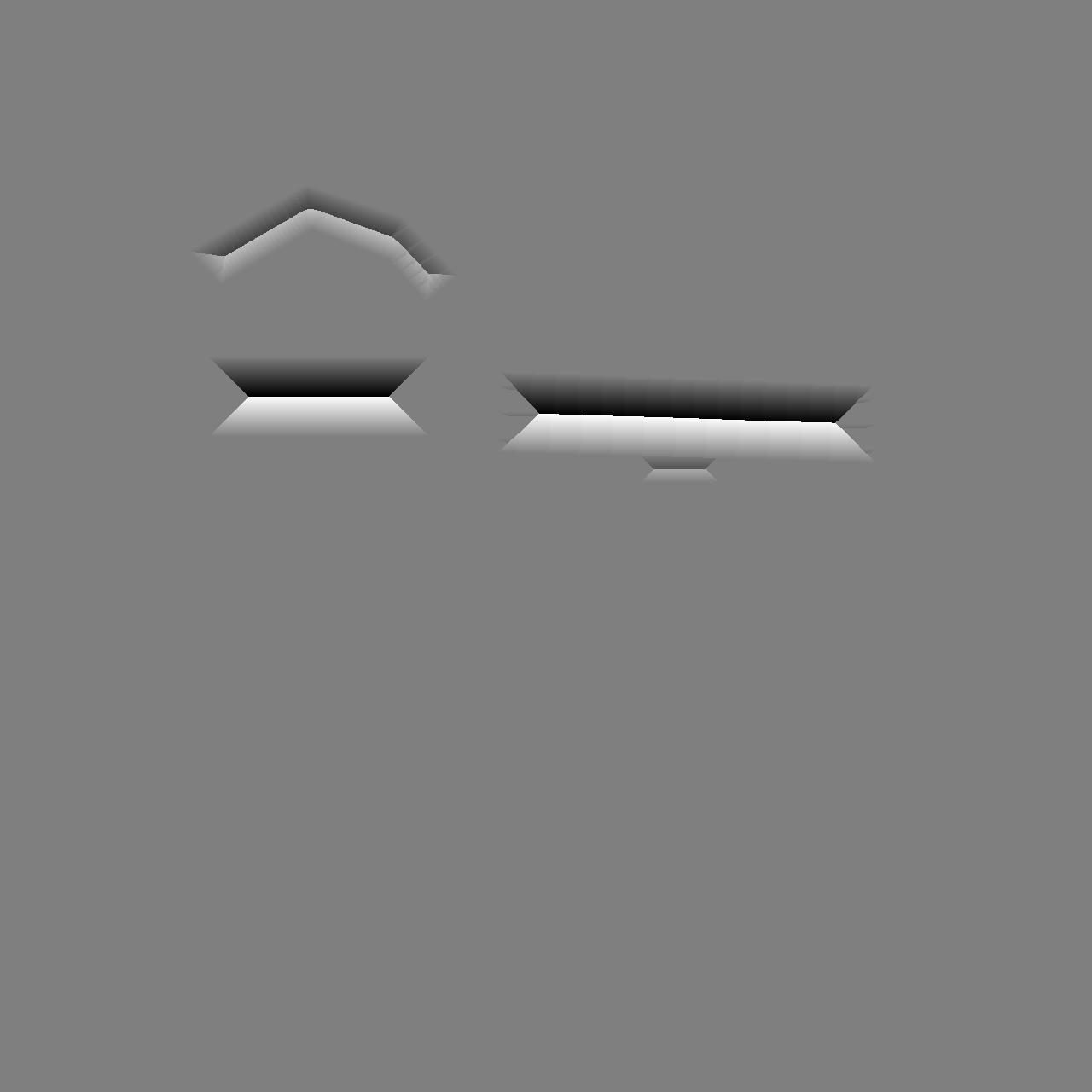}}
\centerline{\small{ (e) X Direction Map}}
\end{minipage}
\vspace{3pt}
\begin{minipage}[c]{0.32\linewidth}
\centerline{\includegraphics[width=\textwidth,height=0.8in]{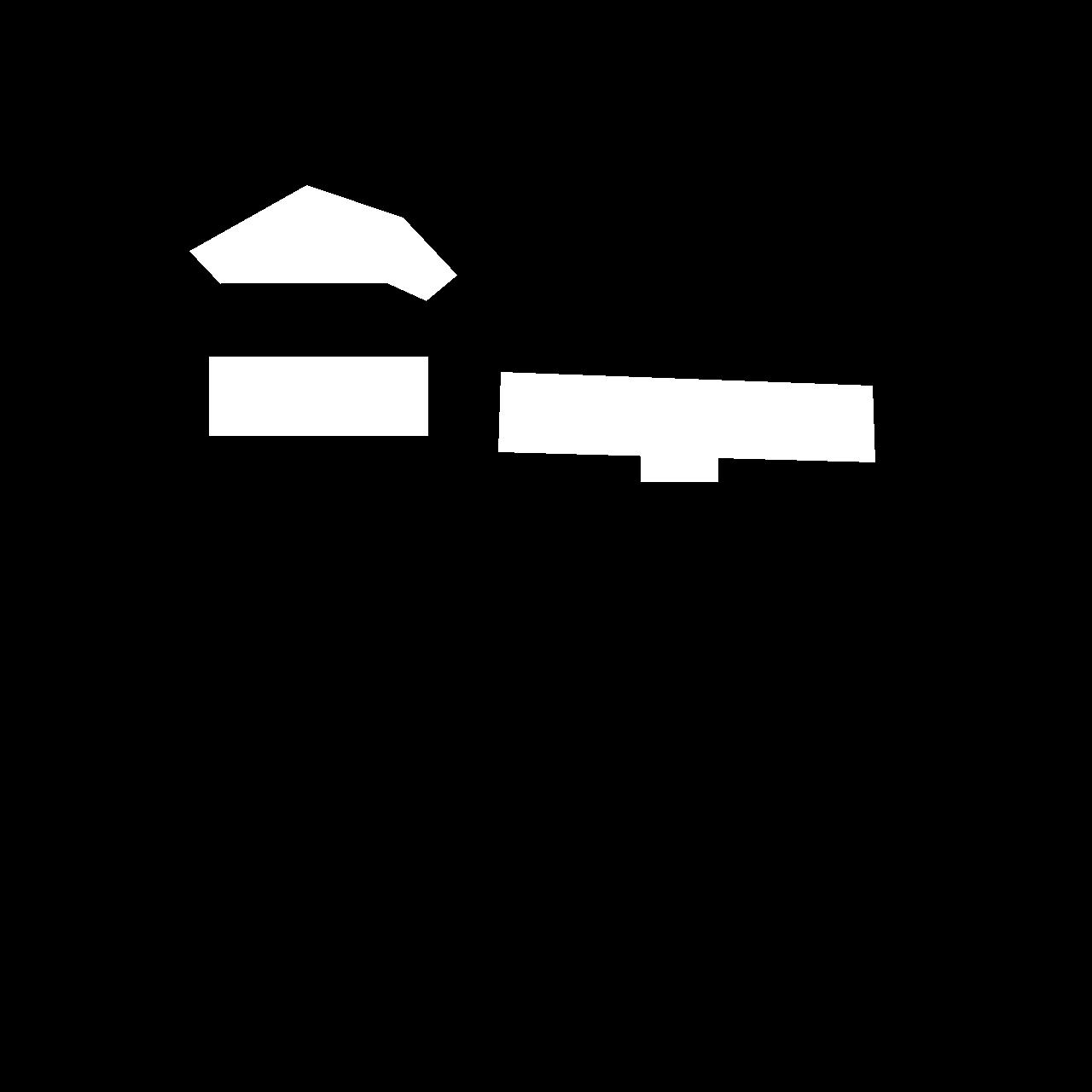}}
\centerline{\small{ (c) Text foreground Map}}
\vspace{3pt}
\centerline{\includegraphics[width=\textwidth,height=0.8in]{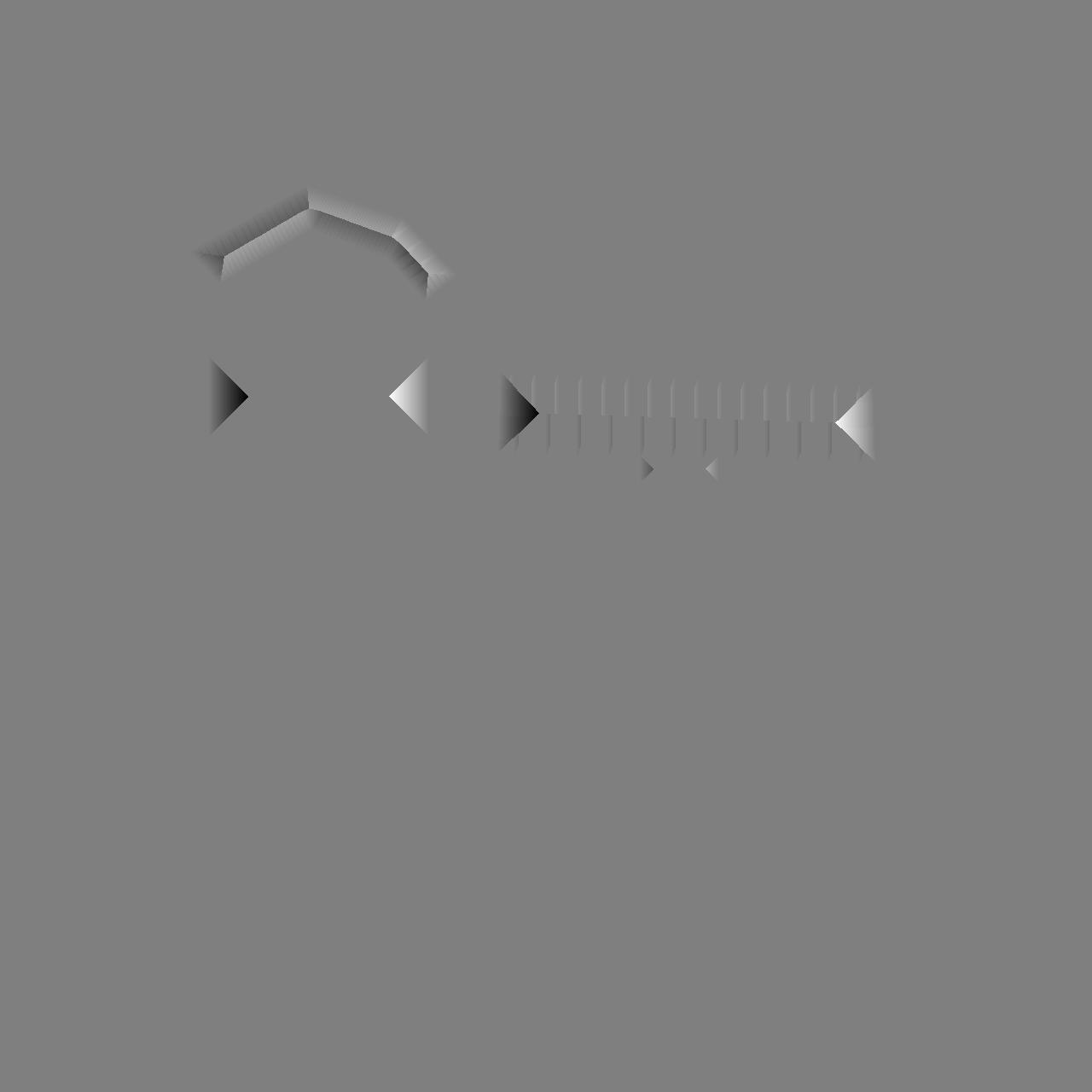}}
\centerline{\small{ (f) Y Direction Map}}
\end{minipage}
\caption{Examples of sample ground-truth labels for the total-text dataset. 'X Direction Map' represents the visualization in the x-direction and 'Y Direction Map' represents the visualization in the y-axis direction.}
\label{Figure3.}
\end{figure}
\end{center}

\subsection{Training Objective Loss}
\label{sec: Training Objective Loss}
\noindent {\bfseries Label generation:} We designed four types of text region information, including text center maps, foreground classification maps, edge maps, and pixel direction maps. Correspondingly, four types of segmentation labels need to be generated. 

To avoid incorrectly segmenting adjacent text regions into a connected domain, we perform segmentation detection on the central region of the text. The center region is obtained by using the Vatti algorithm \cite{11} to clip it from the text foreground region. The foreground classification label of a text region contains the entire text region, with pixel labels inside the region as 1 and outside the region as 0. The distance map is the distance from each pixel inside the text region to the nearest text edge point, and the direction map is the direction from each pixel inside the text region to the nearest edge point. The visual results after normalization of the four types of labels are shown in Fig. \ref{Figure3.}. It should be noted that the direction is divided into x-axis direction and y-axis direction, so its feature map is a two-channel tensor.

\noindent {\bfseries Multitask loss function:} To perceive multiple information about text regions, we designed a multi-objective training function. The objective function is defined as follows:
\begin{equation}
Loss=\alpha_1L_{cen}\times{\alpha_2L}_{for}+{\alpha_3L}_{dis}+{\alpha_4L}_{dir}+{\alpha_5L}_b
\end{equation}
where $L_{cen}$, $L_{for}$, $L_{dis}$ and $L_{dir}$, represent the cross-entropy loss functions between the corresponding predicted probability map $D_{Tar}$ and ground truth labels.$L_{b}$ is Differentiable Binarization loss\cite{3}. $\alpha_1$,$\alpha_2$, $\alpha_3$, $\alpha_4$, $\alpha_5$, and $\alpha_6$ are balancing factors, all set to 1 in the experiment.

\begin{table}[h!]
\caption{Text detection performances comparison with other methods on TotalText. Bold indicates the highest indicator.}
\label{Table1.}
\setlength{\tabcolsep}{2mm}{
\begin{center}
\begin{tabular}{l|c|c c c}
\toprule
{Methods}& {Paper} & R(\%) & P(\%) & F(\%) \\ 
\midrule
TextSnake\cite{26} & ECCV'18 & 74.5 & 82.7 & 78.4\\
PSENET \cite{23}&CVPR'19 & 75.2 & 84.5 & 79.6 \\
CRAFT \cite{24}& CVPR'19 & 79.9 & 87.6 & 83.6 \\
DRRG\cite{22}&CVPR'20 & {\bfseries 84.9} & 86.5 & 85.7 \\
DB  \cite{3} & AAAI'20  & 82.5 & 87.1 & 84.7 \\
FCENet \cite{10}& CVPR'21  & 82.7 & 85.1 & 83.9\\
PCR \cite{20} & CVPR'21 & 82.0 & 88.5 & 85.2 \\
MS-ROCANet \cite{1} & ICASSP'22 &  83.3 & 85.6 & 84.5\\

DB++ \cite{4} & TPAMI'23 & 83.2 & 88.9 & 86.0 \\
TCM-DBNet \cite{29} & CVPR'23 & -&-& 85.9  \\
\hline
RMIPN & Ours & 83.0 & {\bfseries 89.4} &  {\bfseries 86.1}\\
\bottomrule
\end{tabular}
\end{center}
}
\end{table}

\section{Experiments}
\label{sec:Experiments}
In order to verify the effectiveness of our proposed algorithm, we conducted experiments on publicly available benchmark datasets. This section will introduce the experimental details.

\subsection{Datasets}
\label{sec:Datasets}

SynthText dataset \cite{12} contains 800K samples of text rendered on scene images. The synthetic dataset is only used for training.
TotalText \cite{14} is divided into a training set of 1255 images and a testing set of 300 images. The scene text in this dataset has features such as arbitrary shape and varying orientation. The text in the scene has variable numbers of edge point annotations.
MSRA-TD500 \cite{16} consists of a training subset of 300 images and a testing subset of 200 images. The dataset contains multiple languages, including English and Chinese. Text instances are annotated with text lines.  Consistent with previous research \cite{3}, we extracted 400 images from HUST-TR400 \cite{17} for training.

\subsection{Implementation Details}
\label{sec: Implementation Details}
In fact, it is difficult to make an absolutely fair comparison between different algorithms due to the differences in backbones and other training strategies. In order to make the comparison as fair as possible, we adopted similar or the same training strategies and settings with previous algorithms \cite{3,4}. The algorithm code is implemented in PyTorch and trained on a single NVIDIA TITAN RTX graphics card with 24G. The model was initialized using the DB++ \cite{4} provided by their paper. During training, the same data enhancement techniques are used as in DB \cite{3}, DB++\cite{4}, including random rotation, cropping, flipping, etc. During the training process, pre-training is first conducted on the SynthText for 50k iterations, followed by fine-tuning the corresponding real dataset for 1000 epochs. We used the SGD optimizer with a batch size of 2. The learning rate is initialized to 0.001, the weight decay is set to 0.0001, and the momentum is set to 0.9.

\subsection{Comparison with SOTA approaches}
\label{sec:Comparison}
\begin{center}
\begin{figure}[h]
\begin{minipage}[c]{0.32\linewidth}
\centerline{\includegraphics[width=\textwidth,height=0.7in]{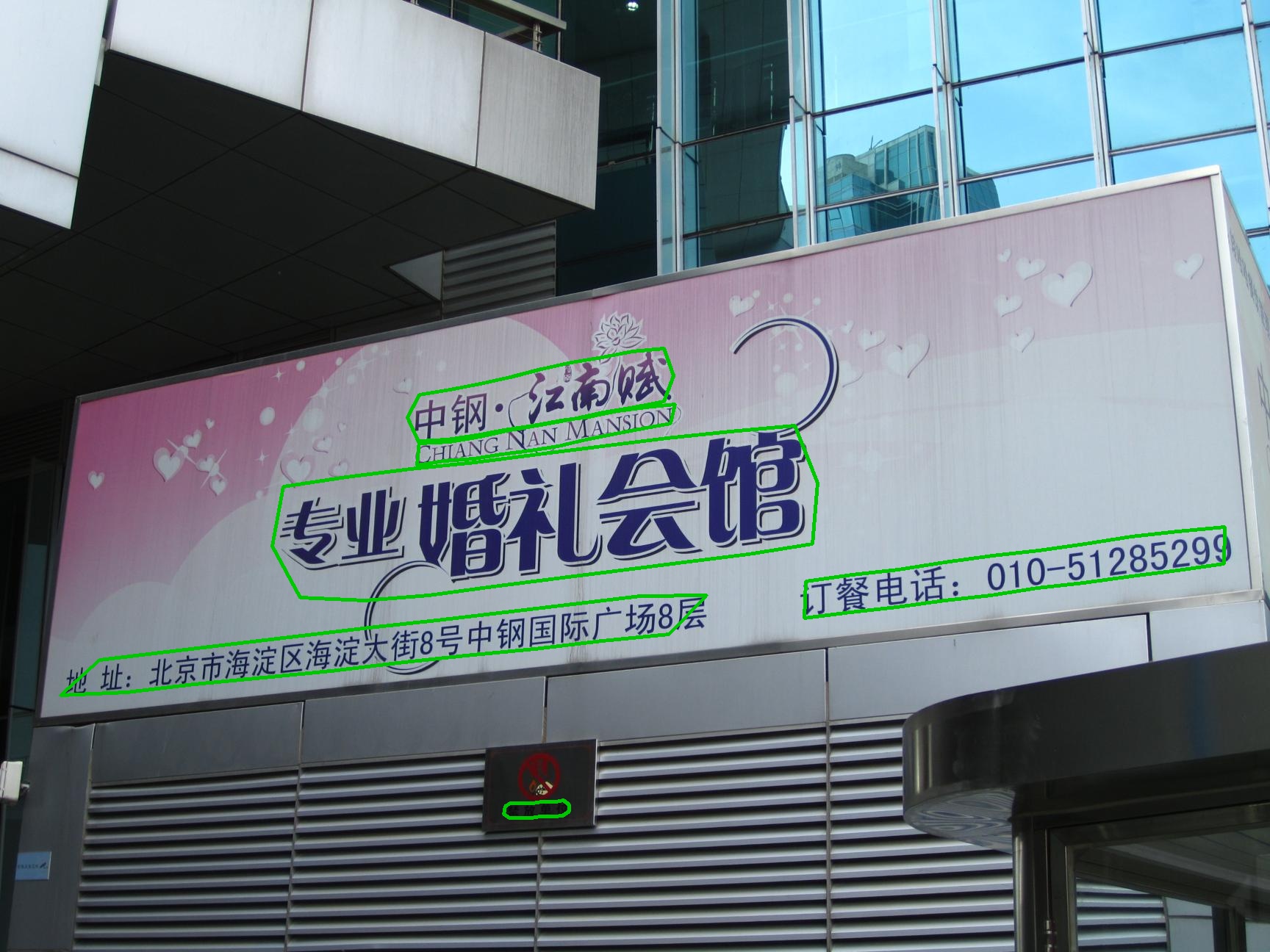}}
\end{minipage}
\vspace{3pt}
\begin{minipage}[c]{0.32\linewidth}
\centerline{\includegraphics[width=\textwidth,height=0.7in]{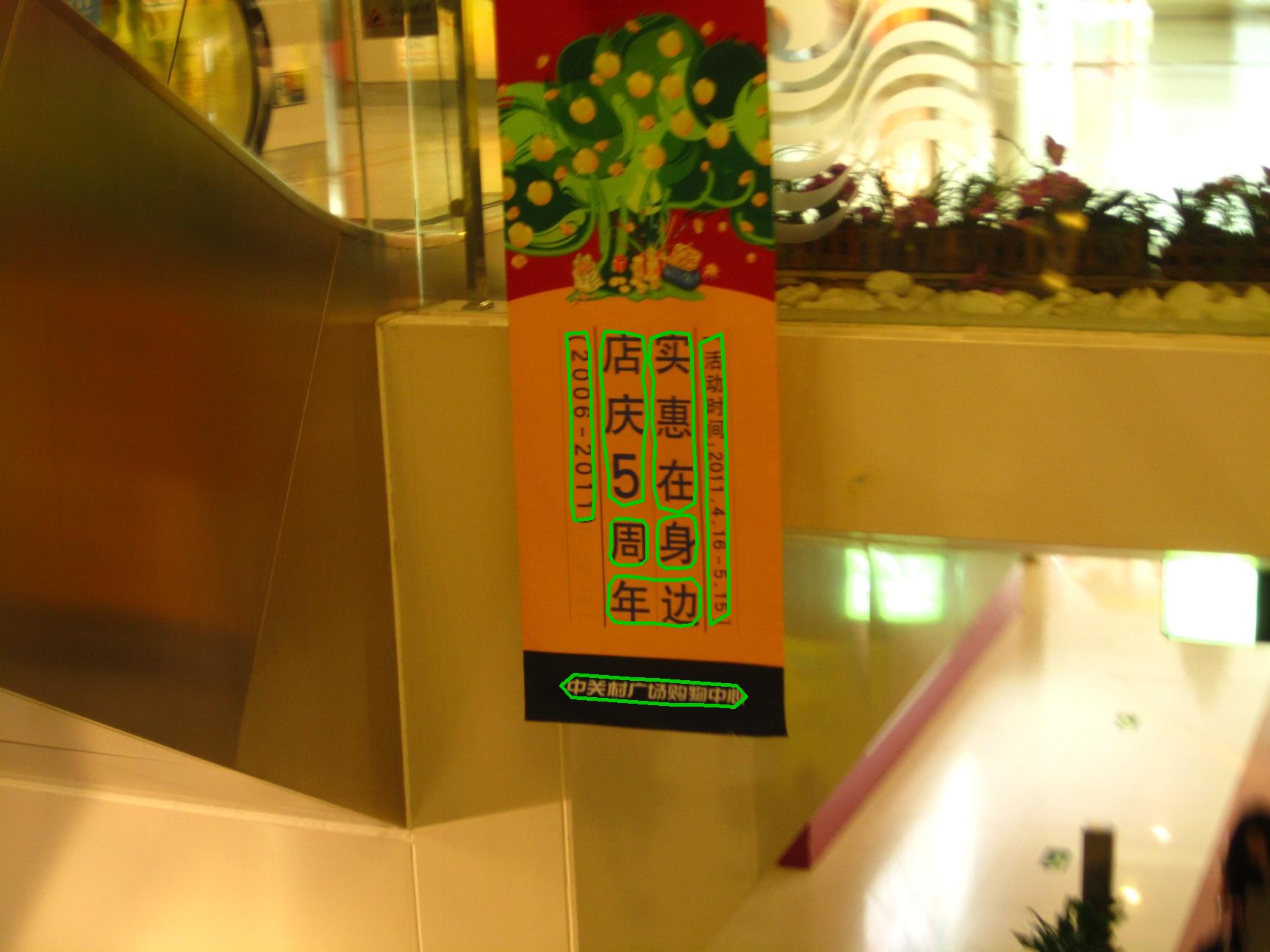}}
\end{minipage}
\vspace{3pt}
\begin{minipage}[c]{0.32\linewidth}
\centerline{\includegraphics[width=\textwidth,height=0.7in]{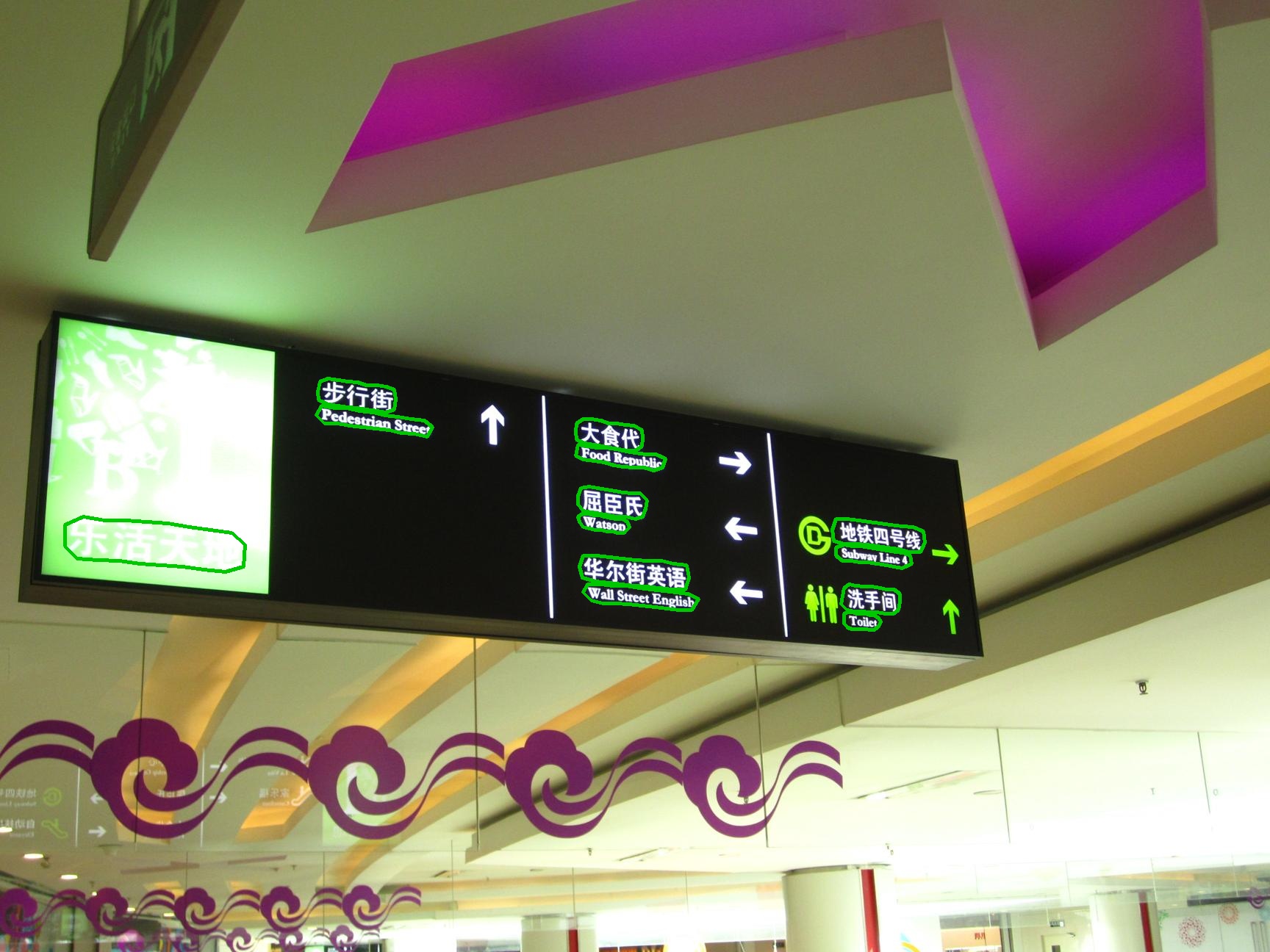}}
\end{minipage}

\vspace{3pt}
\centerline{\small{ (a) MSRA-TD500}}

\begin{minipage}[c]{0.32\linewidth}
\centerline{\includegraphics[width=\textwidth,height=0.7in]{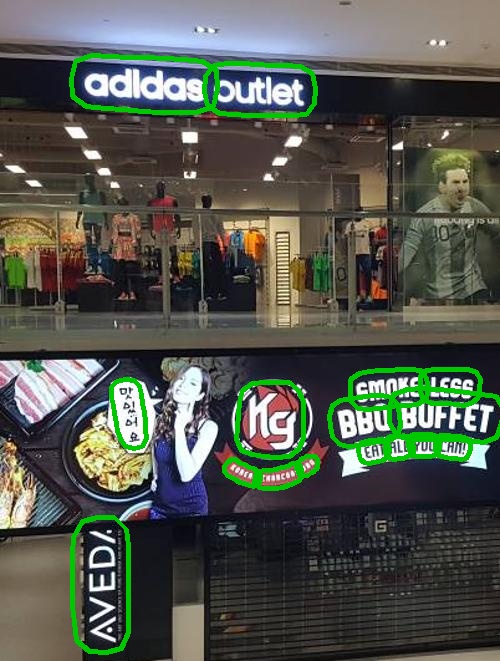}}
\end{minipage}
\vspace{3pt}
\begin{minipage}[c]{0.32\linewidth}
\centerline{\includegraphics[width=\textwidth,height=0.7in]{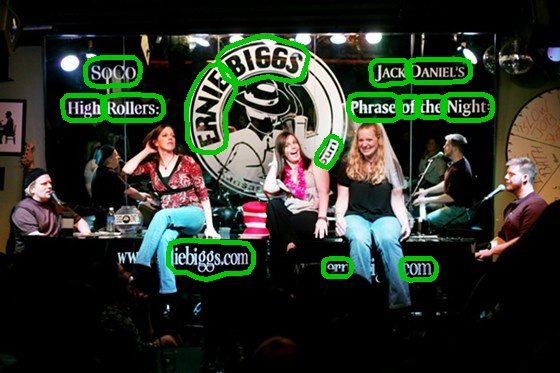}}
\end{minipage}
\vspace{3pt}
\begin{minipage}[c]{0.32\linewidth}
\centerline{\includegraphics[width=\textwidth,height=0.7in]{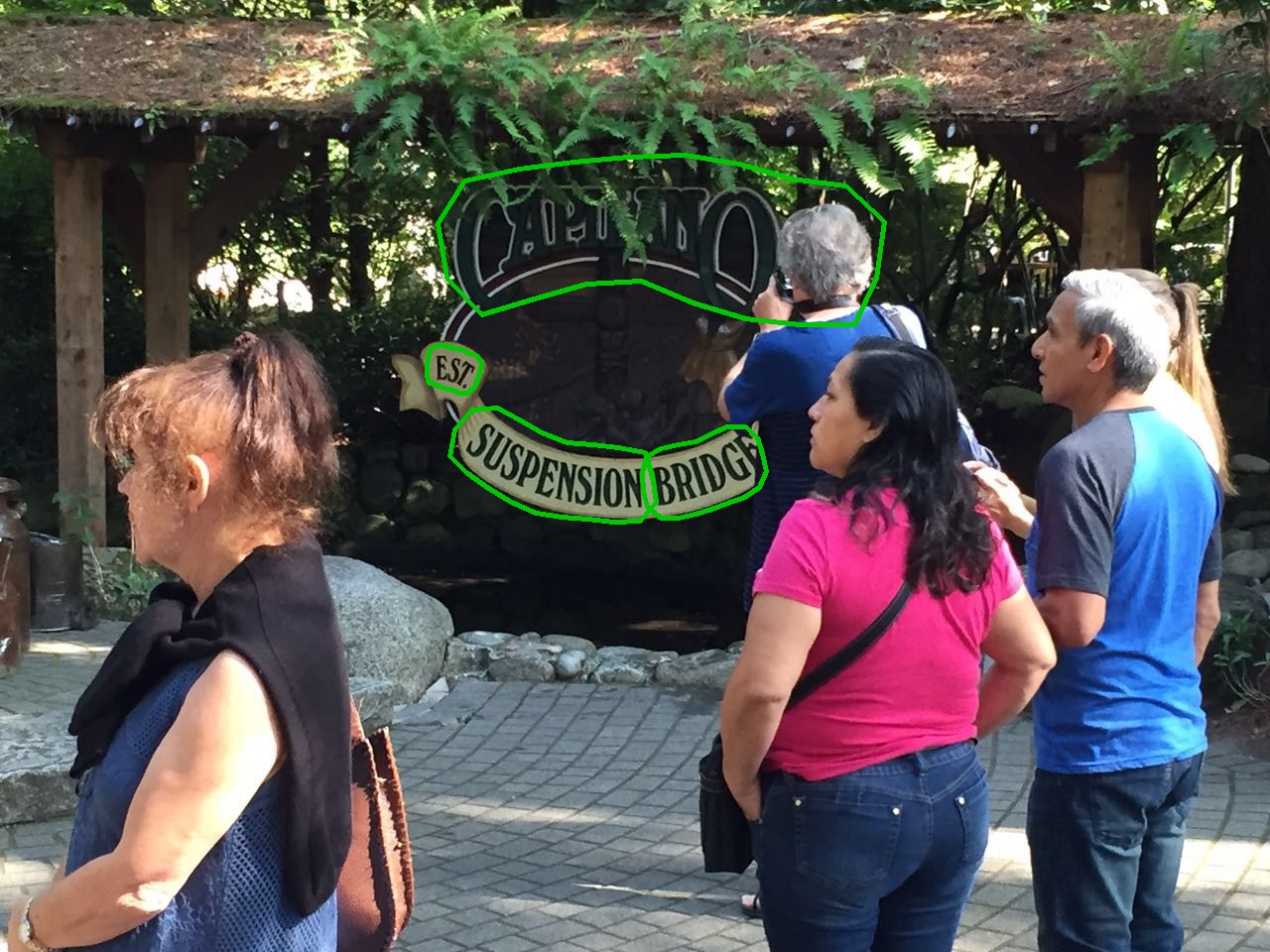}}
\end{minipage}
\centerline{\small{ (c) Total-Text}}

\caption{Example of RMIPN detection results on datasets MSRA-TD500, ICDAR2015, and TotalText.(a) MSRA-TD500 dataset, (b) TotalText data set}
\label{Figure4.}
\end{figure}
\end{center}

We compared our proposed algorithm with the state-of-the-art algorithms, and the statistics are shown in Tab.\ref{Table1.}, Tab \ref{Table2.}. It can be seen that our proposed algorithm achieved comparable performance with SOTA algorithms. Specifically, our algorithm outperformed SOTA algorithms on the MSRA-TD500 dataset. The visualized detection results of our proposed algorithm are shown in Fig.\ref{Figure4.}.

\begin{table}[h!]
\caption{Text detection performances comparison with other methods on MSRA-TD500. Bold indicates the highest indicator.}
\label{Table2.}
\setlength{\tabcolsep}{2mm}{
\begin{center}
\begin{tabular}{l|c|c c c}
\toprule
Methods & Paper & R(\%) & P(\%) & F(\%) \\
\midrule
SAE \cite{25}& CVPR'19 & 81.7 & 84.2 & 82.9\\
CRAFT \cite{24}& CVPR'19 & 78.2 & 88.2 & 82.9 \\
DRRG\cite{22}&CVPR'20 & 82.3 & 88.1 & 85.1\\
DB  \cite{3} & AAAI'20  & 79.2& 91.5 & 84.8\\
MOST  \cite{19}& CVPR'21 & 82.7 & 90.4 & 86.4\\
PCR \cite{20} & CVPR'21 & 83.5 & 90.8 & 87.0\\
FC$^2$RN \cite{30} & ICASSP'21 & 81.8 & 90.3 & 85.8 \\
DB++ \cite{4} & TPAMI'23 & 83.3 & 91.5 & 87.2\\
TCM-FCENet \cite{29} & CVPR'23 & - & -& 86.9 \\
\hline
MIPN & Ours &  {\bfseries 83.4} & {\bfseries 92.1} & {\bfseries 87.5}\\
\bottomrule
\end{tabular}
\end{center}
}
\end{table}

\subsection{Ablation Study}
\label{sec: Ablation Study}
To validate the effectiveness of RMIPM as a plug-and-play module, we set up an ablation study on the MSRA-TD500. 
The statistical results are shown in Tab \ref{Table3.}.
By inserting the RMIPM model into the DB algorithm, recall rate, accuracy, and F-value can be improved to some extent. This proves the effectiveness of RMIPM.

\begin{table}[h]
\caption{Ablation study of RMIPM on the two datasets. DB+RMIPM indicates that RMIPM is embedded into the DB.}

\setlength{\tabcolsep}{0.5mm}{
\begin{center}
\begin{tabular}{l | c  c  c | c c c}
\toprule
\multirow{2}{*}{Methods}&\multicolumn{3}{c|}{MSRA-TD500} & \multicolumn{3}{c}{Total-Text} \\ 
\cline{2-7}
\multirow{1}{*}& R(\%) & P(\%) & F(\%) & R(\%) & P(\%) & F(\%)  \\ 
\midrule
DB \cite{3} & 79.2 & 91.5 & 84.9 & 82.5 & 87.1 & 84.7  \\
DB+RMIPM &  {\bfseries 80.8} &  {\bfseries 91.8 }&  {\bfseries 86.0} &  {\bfseries 82.6} &  {\bfseries 88.1 }& {\bfseries 84.8} \\
\bottomrule
\end{tabular}
\end{center}
}
\label{Table3.}
\end{table}

\subsection{CONCLUSION}
\label{sec:CONCLUSION}
To improve the detection accuracy of existing segmentation-based scene text detection algorithms by perceiving multiple text region information, we proposed a Region Multiple Information Perception Network(RMIPN). The network consists of three parts: a backbone, a RMIPM, and a  detection head. RMIPM is composed of multiple IPMs, each of which can perceive the individual information of text regions based on perceived targets. In the paper, a text center region information, a text foreground classification map, a text pixel distance map, and a text pixel direction map are designed as four types of perceivable text region information. Experiments on public benchmarks have shown that our proposed algorithm can improve text detection performance and is effective. Therefore, the work of this paper provides an idea for the development of segmentation-based scene text detection algorithms. In the future, we will explore building detection algorithms with knowledge reasoning capabilities between scene texts to promote the development of scene text understanding tasks.

\bibliographystyle{IEEEbib}

\end{document}